\documentclass[10pt,journal,compsoc]{IEEEtran}
\ifCLASSOPTIONcompsoc
  \usepackage[nocompress]{cite}
\else
  \usepackage{cite}
\fi
\ifCLASSINFOpdf
\else
\fi
\usepackage{microtype}
\usepackage{graphicx}
\usepackage{subfigure}
\usepackage{booktabs} % for professional tables
\usepackage{tabularx}
\usepackage[table]{xcolor}
\usepackage{amsmath,amssymb}

\usepackage{hyperref}

\usepackage{siunitx}
\usepackage{multirow}

\usepackage[utf8]{inputenc}
\usepackage{fancyhdr}

\fancypagestyle{firstpage}{%
  \chead{This work has been submitted to the IEEE for possible publication. Copyright may be transferred without notice, after which this version may no longer be accessible.}
}

\usepackage{xcolor}

\newcommand{\E}{\mathbb{E}}
\newcommand{\Var}{\mathbb{V}ar}
\newcommand{\Cov}{\mathbb{C}ov}

\begin{document}
%
% paper title
% Titles are generally capitalized except for words such as a, an, and, as,
% at, but, by, for, in, nor, of, on, or, the, to and up, which are usually
% not capitalized unless they are the first or last word of the title.
% Linebreaks \\ can be used within to get better formatting as desired.
% Do not put math or special symbols in the title.
\title{Fast Approximate Multi-output Gaussian Processes}
%
%
% author names and IEEE memberships
% note positions of commas and nonbreaking spaces ( ~ ) LaTeX will not break
% a structure at a ~ so this keeps an author's name from being broken across
% two lines.
% use \thanks{} to gain access to the first footnote area
% a separate \thanks must be used for each paragraph as LaTeX2e's \thanks
% was not built to handle multiple paragraphs
%
%
%\IEEEcompsocitemizethanks is a special \thanks that produces the bulleted
% lists the Computer Society journals use for "first footnote" author
% affiliations. Use \IEEEcompsocthanksitem which works much like \item
% for each affiliation group. When not in compsoc mode,
% \IEEEcompsocitemizethanks becomes like \thanks and
% \IEEEcompsocthanksitem becomes a line break with idention. This
% facilitates dual compilation, although admittedly the differences in the
% desired content of \author between the different types of papers makes a
% one-size-fits-all approach a daunting prospect. For instance, compsoc 
% journal papers have the author affiliations above the "Manuscript
% received ..."  text while in non-compsoc journals this is reversed. Sigh.

\author{Vladimir Joukov and Dana Kulić
\IEEEcompsocitemizethanks{ \IEEEcompsocthanksitem Vladimir Joukov is at University of Waterloo, Canada. \protect\\
E-mail: vjoukov@uwaterloo.ca
\IEEEcompsocthanksitem Dana Kuli\'{c} is with the Faculty of Engineering, Monash University, Melbourne,
Australia.\protect\\
% note need leading \protect in front of \\ to get a newline within \thanks as
% \\ is fragile and will error, could use \hfil\break instead.
E-mail: Dana.Kulic@monash.edu}% <-this % stops an unwanted space
}

\IEEEtitleabstractindextext{%

\begin{abstract}
Gaussian processes regression models are an appealing machine learning method as they learn expressive non-linear models from exemplar data with minimal parameter tuning and estimate both the mean and covariance of unseen points. However, exponential computational complexity growth with the number of training samples has been a long standing challenge. During training, one has to compute and invert an $N \times N$ kernel matrix at every iteration. Regression requires computation of an $m \times N$ kernel where $N$ and $m$ are the number of training and test points respectively.  In this work we show how approximating the covariance kernel using eigenvalues and functions leads to an approximate Gaussian process with significant reduction in training and regression complexity. Training with the proposed approach requires computing only a $N \times n$ eigenfunction matrix and a $n \times n$ inverse where $n$ is a selected number of eigenvalues. Furthermore, regression now only requires an $m \times n$ matrix. Finally, in a special case the hyperparameter optimization is completely independent form the number of training samples. The proposed method can regress over multiple outputs, estimate the  derivative of the regressor of any order, and learn the correlations between them. The computational complexity reduction, regression capabilities, and multioutput correlation learning are demonstrated in simulation examples. 
\end{abstract}

% Note that keywords are not normally used for peerreview papers.
}

% make the title area
\maketitle
\thispagestyle{firstpage}

% To allow for easy dual compilation without having to reenter the
% abstract/keywords data, the \IEEEtitleabstractindextext text will
% not be used in maketitle, but will appear (i.e., to be "transported")
% here as \IEEEdisplaynontitleabstractindextext when the compsoc 
% or transmag modes are not selected <OR> if conference mode is selected 
% - because all conference papers position the abstract like regular
% papers do.
\IEEEdisplaynontitleabstractindextext
% \IEEEdisplaynontitleabstractindextext has no effect when using
% compsoc or transmag under a non-conference mode.

% For peer review papers, you can put extra information on the cover
% page as needed:
% \ifCLASSOPTIONpeerreview
% \begin{center} \bfseries EDICS Category: 3-BBND \end{center}
% \fi
%
% For peerreview papers, this IEEEtran command inserts a page break and
% creates the second title. It will be ignored for other modes.
\IEEEpeerreviewmaketitle

%\section{Introduction}

\section{Introduction}
\label{background}

Gaussian Processes (GPs) are a non-parametric function and covariance approximation method. Formally, a GP is defined as ``collection of random variables any finite number of which have a joint Gaussian distribution'' \cite{rasmussen2003gaussian}. They have excellent regression capabilities, providing a non-parametric, highly non-linear models. Furthermore, due to their probabilistic nature GPs allow estimating the uncertainty at the output. They have been used extensively in many applications, including geostatistics \cite{lantuejoul2013geostatistical}, robotic modeling and control \cite{deisenroth2013gaussian, nguyen2009model}, and finance \cite{gonzalvez2019financial}. However, the computational complexity of both learning a GP model and utilizing it for regression grows exponentially with the number of training data samples. This has limited their application to smaller data sets.

A GP $f(x) \sim GP(m(x),k(x,x'))$ is completely specified by its mean and covariance functions, $m ( x ) $ and $ k(x,x') $ respectively.
\begin{align}
    m(x) &= \E f(x) \\
    k(x,x^{'} ) &= \E (f(x)-m(x))(f(x')-m(x'))^T)
\end{align}
Notice that the covariance of $f(x)$ is dependent only on the input $x$ and is calculated using a kernel function $k(x,x')$. A kernel is any function that is symmetric and positive definite, leading to a valid positive symmetric definite GP covariance for any input $x$. %Possible kernel functions are discussed in section \ref{sec:kernels}.  
We write the GP as $f(x) \sim GP(m(x),k(x,x'))$, the random variables are thus the value of $f$ at location $x$. 
\iffalse
Since a GP is a collection of random variables, consistency is implied. Consider a jointly Gaussian distribution 
\begin{align}
    \begin{bmatrix}
    y_1 \\
    y_2
    \end{bmatrix} \sim
    \mathcal{N}\left( 
    \begin{matrix}
    \mu_{y_1} \\
    \mu_{y_2}
    \end{matrix},
    \begin{matrix}
    \Sigma_{y_1y_1} & \Sigma_{y_1y_2} \\
    \Sigma_{y_2y_1} & \Sigma_{y_2y_2}
    \end{matrix}\right)
\end{align}
then the marginal distribution of $y_1$ is
\begin{align}
y_1 \sim \mathcal{N}(\mu_{y_1},\Sigma_{y_1y_1})
\end{align}
and the conditional distribution of $(y_1 | y_2)$ is 
\begin{align}
y_1|y_2 \sim \mathcal{N}(\mu_{y_1} + \Sigma_{y_1y_2}\Sigma_{y_2y_2}^{-1}(y_2-\mu_{y_2}),\\ \Sigma_{y_1y_1}-\Sigma_{y_1y_2}\Sigma_{y_2y_2}^{-1}\Sigma_{y_2y_1})
\end{align}

We are typically interested in incorporating training data as prior information. \fi

Consider a set of $N$ observations experiencing zero mean Gaussian noise ${(x_i,y_i)|i \in 1,2\dots N}$ where $y_i = f(x_i) + \epsilon$, with $\epsilon \sim \mathcal{N}(0,\sigma_n^2)$. Assuming a zero mean function $\mu(x) = 0$ and since the noise is independent, $\mathbf{Y} \sim \mathcal{N}(0,K_{\mathbf{X}\mathbf{X}'} + \Sigma_N)$ where $\mathbf{Y} = [y_1, y_2 \dots y_N]^T$, $\mathbf{X} = [x_1, x_2 \dots x_N]^T$, $K_{\mathbf{X}\mathbf{X}'}$ is the kernel matrix $k(\mathbf{X},\mathbf{X}')$, and $\Sigma_N = \sigma_n^2 \mathbf{I}_N$, $\mathbf{I}_N$ being an $N\times N$ identity matrix. Consider now $m$ previously unseen test points $\mathbf{X}_* = [x_1,x_2\dots x_m]^T$, we can condition the prediction using the training data.
\begin{align}
    \mathbf{Y}_* &|\mathbf{X},\mathbf{Y},\mathbf{X}_* \sim \mathcal{N}(\mathbf{\mu}_*,\mathbf{\Sigma}_*)  \\
    \nonumber
    \mathbf{\mu}_* &= \E(\mathbf{Y}_*|\mathbf{X},\mathbf{Y},\mathbf{X}_*) \\ 
    \label{eq:mean_pred}
    &= K_{\mathbf{X}_*\mathbf{X}}(K_{\mathbf{X}\mathbf{X}}+\Sigma_N)^{-1}\mathbf{Y} \\
    \label{eq:cov_pred}
    \mathbf{\Sigma}_* &= \underbrace{K_{\mathbf{X}_*\mathbf{X}_*}}_{m\times m}- \underbrace{K_{\mathbf{X}_*\mathbf{X}}}_{m \times N}\underbrace{(K_{\mathbf{X}\mathbf{X}}+\Sigma_N)^{-1}}_{N \times N} \underbrace{K_{\mathbf{X}\mathbf{X}_*}}_{N \times m}
\end{align}

The size of the training datset $N$ and the size of the test point vector $m$ determine the computational requirements during inference.  While the $N \times N$ matrix $(K_{\mathbf{X}\mathbf{X}}+\Sigma_N)^{-1}$ is constant given training data, the $m \times N$ matrix $K_{\mathbf{X}_*\mathbf{X}}$ must be computed and multiplied with the $N \times N$ matrix to predict $m$ points.

Typically the chosen kernel function $K_{\mathbf{X}\mathbf{X}'}$ will have multiple tuning parameters $\mathbf{\theta}$. A common way to find the optimal parameters $\mathbf{\theta}_*$ for the given training data is to maximize the marginal likelihood. Consider the log likelihood of the training data, assuming that the noise free mean is correct. 
\begin{align}
    \nonumber
    log(P(\mathbf{Y}|\mathbf{X})) &= -\frac{1}{2}\mathbf{Y}^T(K_{\mathbf{X}\mathbf{X}}+\Sigma_N)^{-1}\mathbf{Y} \\
    &-\frac{1}{2}log(|K_{\mathbf{X}\mathbf{X}}+\Sigma_N|) - \frac{N}{2}log(2\pi)
\end{align}
Letting $\mathbf{K} = K_{\mathbf{X}\mathbf{X}}+\Sigma_N$ and differentiating the log likelihood with respect to the $j_{th}$ tuning parameter, gradient descent methods can be utilized to iteratively find $\mathbf{\theta}_*$.
\begin{align}
    \frac{\partial log(P(\mathbf{Y}|\mathbf{X}))}{\partial \theta_j} = \frac{1}{2}\mathbf{Y}^T\mathbf{K}^{-1}\frac{\partial \mathbf{K}}{\partial \theta_j}\mathbf{K}^{-1}\mathbf{Y} -\frac{1}{2}tr(\mathbf{K}^{-1}\frac{\partial \mathbf{K}}{\partial \theta_j})
\end{align}
Notice that at each iteration of gradient descent the $N \times N$ matrix $\mathbf{K}$ needs to be computed and then inverted leading to $\mathcal{O}(N^3)$ operations. This makes GPs limited to smaller training datasets. 

\subsection{Related Work}

One way to reduce the computational complexity is to assume that the entire dataset contains redundant information and thus the GP can be accurately approximated by choosing a smaller set of $h$ inducing points. This can be viewed as approximating the full kernel covariance matrix with one of lower rank \cite{quinonero2005unifying}. Multiple inducing point approximations have been proposed \cite{smola2001sparse, seeger2003fast, snelson2006sparse}. Minimizing the Kullback Leibler divergence between the approximate and full posterior processes allows to optimize both the selection of the inducing points and the kernel hyperparameters \cite{titsias2009variational}. Selecting different inducing points for the mean and covariance estimation (decoupling basis) allows to model more complex mean functions while maintaining computationally tractable covariance \cite{cheng2017variational}. Setting the basis of the mean to contain the basis of the covariance and an additional orthogonal component ensures that the components can be optimized separately \cite{salimbeni2018orthogonally}. Similarly, the full GP can be thought of as combination of two independent processes with inducing points, the variation not captured by the inducing points of the first is contained in the second \cite{shi2020sparse}.

Instead of choosing the inducing points, other approaches to reduce computational complexity focus on approximations of the kernel function. It is possible to approximate certain kernels as an output from a linear, time invariant, stochastic system of finite order \cite{hartikainen2010kalman}. In this form, the required numerical problems deal with symmetric block-tridiagonal matrices and can utilize parallelization to further speed up computation \cite{grigorievskiy2017parallelizable}. Combining both the inducing points and state space approximations leads to "double sparse" GPs, further decreasing the complexity and storage requirements \cite{adam2020doubly}. Similar to the state space approximation one can approximate any kernel as a finite Fourier series and optimize over both the selected frequencies and their coefficients \cite{lazaro2010sparse}.

In this work we show how approximating a covariance kernel using eigenfunctions and eigenvalues can greatly reduce the computational complexity of GP regression and training. Section \ref{sec:famgp} first shows how the eigen decomposition reduces GP regression computational complexity, and that it also leads to a differentiable approximate GP where derivatives of arbitrary order are easily computed. Finally it discusses how using the approximation also reduces the complexity of learning the kernel hyperparameters. Sections \ref{sec:mout} and \ref{sec:kernels} extend the approach to multioutput processes and provide the available kernels and their eigenvalue decompositions respectively. We validate the proposed approach in section \ref{sec:exp} showing the computational complexity, regression accuracy, and its ability to learn correlations between outputs and finally discuss future research directions in section \ref{sec:fut}.

\section{Approximate Kernel Gaussian Processes} \label{sec:famgp}
In this section we first show how approximating the kernel matrix using $n$ eigen functions and values leads to an approximate Gaussian process where the necessary matrix inversion is reduced from a $N \times N$ to $n \times n$. Next, we demonstrate that taking the derivative of the eigen functions also allows for estimating the $k_{th}$ order derivative of the approximate GP. Finally, we investigate the optimization of kernel hyperparameters using gradient descent and show that computational complexity grows linearly with the number of training points as opposed to exponentially in regular GP formulation. Furthermore, in the special case when hyper parameters are present only in eigenvalues, the optimization is independent from the number of training points.

Mercer's theorem states that for any continuous symmetric non-negative definite kernel there exists an orthonormal basis consisting of eigen functions $\Phi_i(x)$ and non-increasing eigen values $\lambda_i$ \cite{minh2006mercer} such that 
\begin{align}
    k(x,x') = \sum_{n=1}^{\infty} \lambda_i \phi_i(x) \phi_i(x')
\end{align}
Let us assume that we know this decomposition for our desired kernel, we can thus approximate $K_{\mathbf{X}\mathbf{X}'}$ by utilizing only $n$ eigen values. In vector notation 
\begin{align}
  K_{\mathbf{X}\mathbf{X}} \approx \mathbf{\Phi}_{\mathbf{X}}\Lambda\mathbf{\Phi_{X'}}^T   
\end{align}
where $\mathbf{\Phi}_{\mathbf{X}i,j} = \phi_j(x_i)|j \in 1\dots n$ and $\Lambda$ is a diagonal matrix of the eigenvlaues $[\lambda_1,\lambda_2\dots \lambda_n ]$. Substituting this approximation into the prediction equations \ref{eq:mean_pred} and \ref{eq:cov_pred},
\begin{align}
\label{eq:mean_approx_pred}
    \mu_* &\approx \mathbf{\Phi}_{\mathbf{X}_*}\Lambda\mathbf{\Phi}_{\mathbf{X}}^T(\mathbf{\Phi}_{\mathbf{X}}\Lambda\mathbf{\Phi}_{\mathbf{X}}^T + \Sigma_N)^{-1}\mathbf{Y} \\
    \label{eq:cov_approx_pred}
    \mathbf{\Sigma}_* &\approx \mathbf{\Phi}_{\mathbf{X}_*}\Lambda\mathbf{\Phi}_{\mathbf{X}_*}^T \\
    &-\mathbf{\Phi}_{\mathbf{X}_*}  \Lambda\mathbf{\Phi}_{\mathbf{X}}^T(\mathbf{\Phi}_{\mathbf{X}}\Lambda\mathbf{\Phi}_{\mathbf{X}}^T + \Sigma_N)^{-1}  
    \mathbf{\Phi}_{\mathbf{X}}\Lambda\mathbf{\Phi}_{\mathbf{X}_*}^T
\end{align}
Recall the binomial inverse theorem 
\begin{align}
    (\mathbf{A} + \mathbf{U}\mathbf{B}\mathbf{V})^{-1} = \mathbf{A}^{-1}-\mathbf{A}^{-1}\mathbf{U}(\mathbf{B}^{-1} + \mathbf{V}\mathbf{A}^{-1}\mathbf{U})^{-1}\mathbf{V}\mathbf{A}^{-1}
\end{align}
which allows us to simplify the inverse of $\mathbf{K_\Phi} = (\mathbf{\Phi}_{\mathbf{X}}\Lambda\mathbf{\Phi}_{\mathbf{X}}^T + \Sigma_N)$ as
\begin{align}
    %(\mathbf{\Phi}_{\mathbf{X}}\Lambda\mathbf{\Phi}_{\mathbf{X}}^T + \Sigma_N)^{-1} = \\ 
    \mathbf{K_\Phi}^{-1} = \Sigma_N^{-1} - 
    \Sigma_N^{-1}\mathbf{\Phi}_{\mathbf{X}}(\Lambda^{-1} &+ \mathbf{\Phi}_{\mathbf{X}}^T\Sigma_N^{-1}\mathbf{\Phi}_{\mathbf{X}})^{-1}\mathbf{\Phi}_{\mathbf{X}}^T\Sigma_N^{-1}
\end{align}

Using this approximation, inference only requires the inverse of an $n \times n$ matrix $\mathbf{\bar{\Lambda}} = \Lambda^{-1}+\mathbf{\Phi}_{\mathbf{X}}^T\Sigma_N^{-1}\mathbf{\Phi}_{\mathbf{X}}$. Substituting this result into the approximate prediction equations \ref{eq:mean_approx_pred} and \ref{eq:cov_approx_pred} leads to significantly faster prediction  compared to regular GP, single output prediction equations are summarized in table \ref{tab:prediction}. 

\begin{table*}[ht]
  \centering
  \caption{Comparison of the proposed FAMGP and regular Gaussian Process regression equations and their respective matrix sizes when predicting the mean $\mu_*$ and covariance $\mathbf{\Sigma}_*$ of the output $\mathbf{Y} \in \mathbb{R}^{m \times 1}$ at $m$ points $\mathbf{x_*} = [x_*^1 \ x_*^2 \ \dotsc x_*^m]^T$.}
    \begin{tabularx}{\textwidth}{|l|X|l|}
    \hline & FAMGP & GP \\ \hline
       Mean & $\begin{aligned}
            \mu_* &= \underbrace{\mathbf{\Phi}_{\mathbf{X}_*}}_{H \times n} \underbrace{\mathbf{\alpha}'}_{n \times 1} \\
            \mathbf{\alpha}' &= \Lambda\mathbf{\Phi}_{\mathbf{X}}^T (
                \Sigma_N^{-1} - 
                \Sigma_N^{-1}\mathbf{\Phi}_{\mathbf{X}}\mathbf{\bar{\Lambda}}^{-1}\mathbf{\Phi}_{\mathbf{X}}^T\Sigma_N^{-1}
                )\mathbf{Y}
        \end{aligned}$ & 
        $\begin{aligned}
        \mathbf{\mu}_* &= \underbrace{K_{\mathbf{X}_*\mathbf{X}}}_{m \times N}
        \underbrace{(K_{\mathbf{X}\mathbf{X}}+\Sigma_N)^{-1}\mathbf{Y}}_{N\times 1}
        \end{aligned}$  \\ \hline
        Covariance &
        $\begin{aligned}
            \mathbf{\Sigma}_* &= \underbrace{\mathbf{\Phi}_{\mathbf{X}_*}}_{m \times n} \underbrace{\mathbf{G}}_{n \times n} \underbrace{\mathbf{\Phi}_{\mathbf{X}_*}^T}_{n \times m} \\
            \mathbf{G} &= \Lambda\mathbf{\Phi}_{\mathbf{X}}^T (
                \Sigma_N^{-1} - 
                \Sigma_N^{-1}\mathbf{\Phi}_{\mathbf{X}}\mathbf{\bar{\Lambda}}^{-1}\mathbf{\Phi}_{\mathbf{X}}^T\Sigma_N^{-1}
                ) \mathbf{\Phi}_{\mathbf{X}} \Lambda
        \end{aligned}$
        & $\begin{aligned}
        \mathbf{\Sigma}_* &= \underbrace{K_{\mathbf{X}_*\mathbf{X}_*}}_{m\times m}- \underbrace{K_{\mathbf{X}_*\mathbf{X}}}_{m \times N}\underbrace{(K_{\mathbf{X}\mathbf{X}}+\Sigma_N)^{-1}}_{N \times N} \underbrace{K_{\mathbf{X}\mathbf{X}_*}}_{N \times m}
        \end{aligned}$  \\ \hline
        Terms & 
                $\begin{array}{l}
                    \mathbf{\Phi}_{\mathbf{X}_*} \textrm{: Kernel eigen function of prediction points \ } \mathbf{x_*} \\
                    \mathbf{\Phi}_{\mathbf{X}} \textrm{ \ : Kernel eigen function of training points \ } \mathbf{x} \\
                    \Lambda \textrm{ \ \ \ \enskip : Kernel eigen values} \\
                    n \textrm{ \ \ \ \ \ : Number of selected eigen values}\\
                    \Sigma_N \textrm{ \ \ : Training output data noise covariance matrix}\\
                    \mathbf{\bar{\Lambda}} = \Lambda^{-1} + \mathbf{\Phi}_{\mathbf{X}}^T\Sigma_N^{-1}\mathbf{\Phi}_{\mathbf{X}} 
                \end{array}$
        & 
                $\begin{array}{l}
                    K_\mathbf{X_*X} \textrm{: Kernel between prediction points \ } \mathbf{x_*} \textrm{\ and training points \ } \mathbf{x} \\
                    K_\mathbf{XX} \textrm{ \ : Kernel between training points \ } \mathbf{x} \\
                    \mathbf{Y} \textrm{\ \ \ \ \ \ \ : Training outputs \ } \\
                    \Sigma_N \textrm{ \ \ \ \ : Training output data noise covariance matrix}
                \end{array}$
        \\ \hline
    \end{tabularx}
    \label{tab:prediction}
\end{table*}

\subsection{Differentiation} \label{sec:diff}
Since differentiation is a linear operator, the derivative of the GP output with respect to the input is also a GP \cite{mchutchon2013differentiating}. Consider two test points $x_*$ and $x_* + \delta$, the respective outputs are then random variables as follows:
\begin{align}
y_* = \Phi_{x_*}\mathbf{\alpha}' + \epsilon_* \\
y_\delta = \Phi_{x_*+\delta}\mathbf{\alpha}' + \epsilon_\delta
\end{align}
where $\epsilon_*, \ \epsilon_\delta \sim \mathcal{N}(0,\Sigma_N^2)$. The two random variables will have a jointly Gaussian distribution 
\begin{align}
    \begin{bmatrix}
    y_* \\
    y_\delta
    \end{bmatrix} \sim
    \mathcal{N}\left( 
    \begin{matrix}
     \Phi_{x_*}\mathbf{\alpha}' \\
    \Phi_{x_*+\delta}\mathbf{\alpha}'
    \end{matrix}\biggr\lvert
    \begin{matrix}
    \Phi_{x_*}G\Phi_{x_*}^T & \Phi_{x_*}G\Phi_{x_*+\delta}^T \\
    \Phi_{x_*+\delta}G\Phi_{x_*} & \Phi_{x_*+\delta}G\Phi_{x_*+\delta}^T
    \end{matrix}\right)
\end{align}
The derivative is thus 
\begin{align}
    \frac{\partial y_*}{\partial x_*} &= \lim_{\delta\to 0} \frac{\Phi_{x_*+\delta}\mathbf{\alpha}' - \Phi_{x_*}\mathbf{\alpha}'}{\delta} + \lim_{\delta\to 0} \frac{\epsilon_\delta - \epsilon_*}{\delta} \\
    &=\underbrace{\frac{\partial \Phi_*}{\partial x_*}\mathbf{\alpha}'}_{mean} + \underbrace{\lim_{\delta\to 0} \frac{\epsilon_\delta - \epsilon_*}{\delta}}_{variance}
\end{align}
Now we substitute the variance and covariance estimate from the jointly Gaussian distribution for the sum
\begin{align}
    \nonumber
    \Var (\lim_{\delta\to 0} \frac{\epsilon_\delta - \epsilon_*}{\delta}) =& \lim_{\delta\to 0}\frac{1}{\delta^2}\biggr( \Var(\epsilon_\delta) + \Var(\epsilon_*) \\
    \nonumber
    &- \Cov(\epsilon_\delta,\epsilon_*) - \Cov(\epsilon_*,\epsilon_\delta) \biggr) \\
    \nonumber
    = \lim_{\delta\to 0}\frac{1}{\delta^2}\biggr( \Phi_{x_*+\delta}G\Phi_{x_*+\delta}^T &+ \Phi_{x_*}G\Phi_{x_*}^T - \\
    \nonumber \Phi_{x_*+\delta}G\Phi_{x_*} &- \Phi_{x_*}G\Phi_{x_*+\delta}^T \biggr) \\
    = \frac{\partial \Phi_*}{\partial x_*}G\frac{\partial \Phi_*}{\partial x_*}^T \ \ \ \ \ \ \ \ \ \ \ \ \ \  \ \ &
\end{align}
Thus, if $\frac{\partial^k \mathbf{\Phi_{X_*}}}{\partial \mathbf{X_*}^k}$ is known we can compute the mean and variance for the $k_{th}$ derivative. 
\begin{align}
    \frac{\partial^k \mu_*}{\partial \mathbf{X_*}^k} &= \frac{\partial^k \mathbf{\Phi_{X_*}}}{\partial \mathbf{X_*}^k} \mathbf{\alpha}' \\
    \Var (\frac{\partial^k y_*}{\partial \mathbf{X_*}^k}) &= \frac{\partial^k \mathbf{\Phi_{X_*}}}{\partial \mathbf{X_*}^k} \mathbf{G} \frac{\partial^k \mathbf{\Phi_{X_*}}^T}{\partial \mathbf{X_*}^k}
\end{align}
In section  \ref{sec:kernels} we show how the structure of some available eigen functions allows for very fast computation of the derivatives.

\subsection{Hyperparameter Training}
We now consider the gradient required to optimize the hyper parameters of the approximate kernel.
\begin{equation}
    \resizebox{\columnwidth}{!} 
    {$\frac{\partial log(P(\mathbf{Y}|\mathbf{X}))}{\partial \theta_j} = \frac{1}{2}\mathbf{Y}^T\mathbf{K_\Phi}^{-1}\frac{\partial \mathbf{K_\Phi}}{\partial \theta_j}\mathbf{K_\Phi}^{-1}\mathbf{Y} -\frac{1}{2}tr(\mathbf{K_\Phi}^{-1}\frac{\partial \mathbf{K_\Phi}}{\partial \theta_j})$}
\end{equation}
where 
\begin{align}
    \frac{\partial \mathbf{K_\Phi}}{\partial \theta_j} = \frac{\partial \mathbf{\Phi_X}}{\partial \theta_j}\Lambda\mathbf{\Phi_X}^T + \mathbf{\Phi_X}\frac{\partial\Lambda}{\partial \theta_j}\mathbf{\Phi_X}^T + \mathbf{\Phi_X}\Lambda\frac{\partial \mathbf{\Phi_X}^T}{\partial \theta_j}
\end{align}
Thus typical gradient descent hyperparameter optimization would require computing $\mathbf{\Phi_X}$, $\frac{\partial \mathbf{\Phi_{X}}}{\partial \theta_j}$, and the inverse of an $n \times n$ matrix at each iteration, avoiding calculating the full $N \times N$ matrix $\mathbf{K}$ and its inverse. Thus the computational complexity grows linearly with the number of training pairs. Any gradient descent algorithm can be utilized for parameter optimization.

Consider a special case when the hyperparameter $\theta_j$ only appears in the eigen values and not the eigen functions. Then $\mathbf{\Phi_X}$ can be treated as a constant and $\frac{\partial \mathbf{\Phi_{X}}}{\partial \theta_j} = 0$. Using the fact that trace is invariant under cyclic permutations the gradient can be written entirely in terms of $n$ sized matrices and vectors. 
\begin{align}
    \nonumber
    \frac{\partial log(P(\mathbf{Y}|\mathbf{X}))}{\partial \theta_j} =& \frac{1}{2}\mathbf{_Y\Sigma_\Phi}(\frac{\partial\Lambda}{\partial \theta_j}-2\frac{\partial\Lambda}{\partial \theta_j}\mathbf{\bar{\Lambda}}^{-1}\mathbf{_\Phi\Sigma_\Phi} \\ 
    \nonumber
    +& \mathbf{\bar{\Lambda}}^{-1}\mathbf{_\Phi\Sigma_\Phi}\frac{\partial\Lambda}{\partial \theta_j}\mathbf{_\Phi\Sigma_\Phi}  \mathbf{\bar{\Lambda}}^{-1})\mathbf{_Y\Sigma_\Phi}^T\\
    -& tr(\frac{\partial\Lambda}{\partial \theta_j}(\mathbf{I}_n - \mathbf{\bar{\Lambda}}^{-1})\mathbf{_\Phi\Sigma_\Phi})
\end{align}
where $\mathbf{_Y\Sigma_\Phi} = \mathbf{Y^T \Sigma_N^{-1}\Phi_X}$ and $\mathbf{_\Phi\Sigma_\Phi} = \mathbf{\Phi_X^T \Sigma_N^{-1} \Phi_X}$ are constant $1 \times n$ vector and $n \times n$ matrix respectively. Note that $\mathbf{_\Phi\Sigma_\Phi}$ is also present in $\mathbf{\bar{\Lambda}}$. This means that to optimize the hyper parameters that only appear in the eigen values, $\mathbf{\Phi_X^T}$ needs only to be computed once and the iterative convergence process is independent from the number of training data points. As we show in section \ref{sec:kernels}, this is true for various kernel decompositions.

\section{Multioutput Extension} \label{sec:mout}
A simple way to handle multioutput modelling using GPs is to assume that the outputs are independent and train a separate GP for each. However, this approach cannot capture the correlation between different outputs present in the training data. By vectorizing the multioutput training data it is possible to capture cross output correlation \cite{bonilla2008multi}. Consider learning a GP representation of a function with $M$ outputs, provided the training pairs $x_i, [y_i^1 \dots y_i^M]$, re-define the training data as $\mathbf{Y} = [y_1^1 \ y_2^1 \dots y_N^1 \ y_1^2 \dots y_N^2 \dots y_N^M]^T$, vectorizing all of the outputs. We now consider the $NM \times NM$ covariance matrix of $\mathbf{Y}$ 
\begin{align}
\label{eq:mv_cov}
    K_f \otimes \mathbf{K_{XX}} + \Sigma_{NM}
\end{align}
where $K_f$ is an $M \times M$ positive symmetric definite matrix that describes output similarities and $NM \times NM$ matrix $\Sigma_{NM}$ describes the observation noise that now may include covariance between outputs, $\otimes$ denotes the Kronecker product. Note that setting $K_f$ to the identity matrix and keeping $\Sigma_{NM}$ diagonal implies independent outputs similar to training a separate GP for each. Inference can be done for multiple outputs by substitution $K_f \otimes \mathbf{K_{X_*X}}$ for $\mathbf{K_{X_*X}}$. We expand on this method by including the proposed kernel approximation in the multioutput covariance and utilizing Kronecker product properties. 

Substituting the eigenfunction and eigenvalue decomposition and relying on the mixed-product Kronecker product property we can again simplify the covariance inverse. 
\begin{align}
\nonumber
    \mathbf{K_{I\Phi}} &= K_f \otimes (\mathbf{\Phi_X}\Lambda\mathbf{\Phi_X}^T) + \Sigma_{NM} \\
    \nonumber
    \mathbf{K_{I\Phi}}^{-1} &= ((\mathbf{I}_M \otimes \mathbf{\Phi_X}) (K_f \otimes \Lambda) (\mathbf{I}_M \otimes \mathbf{\Phi_X}^T) ) + \Sigma_{NM})^{-1} \\
    \nonumber
    = &\Sigma_{NM}^{-1} - 
    \Sigma_{NM}^{-1}(\mathbf{I}_M \otimes \mathbf{\Phi_X})(K_f^{-1} \otimes \Lambda^{-1} \\ +&(\mathbf{I}_M \otimes \mathbf{\Phi_X}^T)\Sigma_{NM}^{-1} (\mathbf{I}_M \otimes \mathbf{\Phi_X}))^{-1}(\mathbf{I}_M \otimes \mathbf{\Phi_X}^T)\Sigma_{NM}^{-1}
\end{align}
The required inverse is now $nM \times nM$ instead of $NM \times NM$.

Often it is assumed that the observation noise is constant at each sample and thus can be extpressed as $\Sigma_{Nm} = \mathbf{S}_M\otimes \mathbf{I}_N$ where $\mathbf{S}_M$ is an $M \times M$ positive definite matrix. In this case we can further simplify the required $nM \times nM$ matrix inverse into eigen decomposition of smaller matrices and matrix multiplication. Substituting the noise covariance $\mathbf{S}_M\otimes \mathbf{I}_N$ into the inverse, using Kronecker mixed-product property, and following a similar approach to \cite{niati2019inverse} we see that 
\begin{align}
    \label{eq:kron_inv}
    \nonumber
    (K_f^{-1} &\otimes \Lambda^{-1} + (\mathbf{I}_M \otimes \mathbf{\Phi_X}^T)(\mathbf{S}_M\otimes \mathbf{I}_N)^{-1}(\mathbf{I}_M \otimes \mathbf{\Phi_X}))^{-1} \\
    \nonumber
    &= (K_f^{-1} \otimes \Lambda^{-1} + \mathbf{S}^{-1} \otimes \mathbf{\Phi_X}^T\mathbf{\Phi_X})^{-1} \\
    &= (K_f \otimes \Lambda)(\mathbf{S}^{-1}K_f \otimes \mathbf{\Phi_X}^T\mathbf{\Phi_X} \Lambda + \mathbf{I}_M \otimes \mathbf{I}_n)^{-1}
\end{align}
Next we apply eigen decomposition to $\mathbf{S}^{-1}K_f = U_aD_aU_a^{-1}$ and $\mathbf{\Phi_X}^T\mathbf{\Phi_X} \Lambda = U_bD_bU_b^{-1}$ where $U$ denotes the matrix of eigenvectors and $D$ is a diagonal matrix of eigenvalues. Substituting the decomposition back into \ref{eq:kron_inv}, expanding the result using the mixed-product property again, and finally applying the binomial inverse theorem, the inverse simplifies to the following:
\begin{align}
    \label{eq:kron_inv2}
    \nonumber
    &(K_f \otimes \Lambda)(U_aD_aU_a^{-1} \otimes U_bD_bU_b^{-1} + \mathbf{I}_M \otimes \mathbf{I}_n)^{-1} \\
    \nonumber
    &=(K_f \otimes \Lambda)[(U_a \otimes U_b)(D_a \otimes D_b)(U_a^{-1} \otimes U_b^{-1})+ \\
    \nonumber
    & \ \ \ \ \ \ \ \ \mathbf{I}_M \otimes \mathbf{I}_n]^{-1} \\
    \nonumber
    &=(K_f \otimes \Lambda)[\mathbf{I}_M \otimes \mathbf{I}_n \\ 
    \nonumber
    &\ \ \ \ \ \ \ \ -(U_a \otimes U_b)(D_a \otimes D_b +\mathbf{I}_M \otimes \mathbf{I}_n)^{-1}(U_a^{-1} \otimes U_b^{-1})] \\
    \nonumber
    &= (K_f \otimes \Lambda) - \\
    & \ \ \ \ \ \ \ \ (K_f U_a \otimes \Lambda U_b)\underbrace{(D_a \otimes D_b +\mathbf{I}_M \otimes \mathbf{I}_n)^{-1}}_{Diagonal}(U_a^{-1} \otimes U_b^{-1})
\end{align}
In cases of large $M$ and $n$ this approach can significantly decrease the computation time since it avoids the inversion of $nM \times nM$ matrix and instead only requires eigen decomposition of  $n \times n$ and $M \times M$ matrices.

\subsection{Learning $K_f$}
Gradient descent can be utilized to learn the matrix $K_f$ by maximizing marginal log likelihood. To guarantee that $K_f$ remains symmetric positive definite during convergence, it can be parametrized using Cholesky decomposition as $K_f = LL^T$ where $L$ is a lower triangular matrix \cite{bonilla2008multi}. Similar to the special case when hyperparameters only appear in the eigenvalues, the gradient is written entirely in terms of $1 \times nM$ vectors and $nM \times nM$ matrices and only requres an $nM \times nM$ matrix inverse at each optimization iteration. 
\begin{align}
    \nonumber
    \frac{\partial log(P(\mathbf{Y}|\mathbf{X}))}{\partial L} = \frac{1}{2}\mathbf{Y}^T\mathbf{K_{I\Phi}}^{-1}\frac{\partial \mathbf{K_{I\Phi}}}{\partial L}\mathbf{K_{I\Phi}}^{-1}\mathbf{Y} \\ -\frac{1}{2}tr(\mathbf{K_{I\Phi}}^{-1}\frac{\partial \mathbf{K_{I\Phi}}}{\partial L})
\end{align}
where $\frac{\partial \mathbf{K_{I\Phi}}}{\partial L} = (\mathbf{I}_M \otimes \mathbf{\Phi_X}) (\frac{\partial LL^T}{\partial L} \otimes \Lambda) (\mathbf{I}_M \otimes \mathbf{\Phi_X}^T)$ and $\frac{\partial LL^T}{\partial L}$ can be calculated as $\frac{\partial LL^T}{\partial L} = (\mathbf{I}_{(nM)^2} + T)\mathbf{I}_{nM}\otimes L$ where $T$ is a transformation matrix such that $T vec(L) = vec(L^T)$ \cite{wang2013derivatives}.

\section{Available Kernels} \label{sec:kernels}
In this section we present some of the kernels with well known Mercer expansions, their $k_{th}$ order derivatives with respect to the input and gradients with respect to their hyperparameters.  For a more comprehensive list the reader is referred to \cite{fasshauer2015kernel}. Note that in our formulation the scaling of any kernel is handled by the $K_f$ matrix thus we omit the commonly included scaling factors from all of the presented kernels. 

\subsection{Squared Exponential}
\begin{figure}
    \centering
    \includegraphics[trim=5cm 1cm 5cm 0,clip,width=0.5\textwidth]{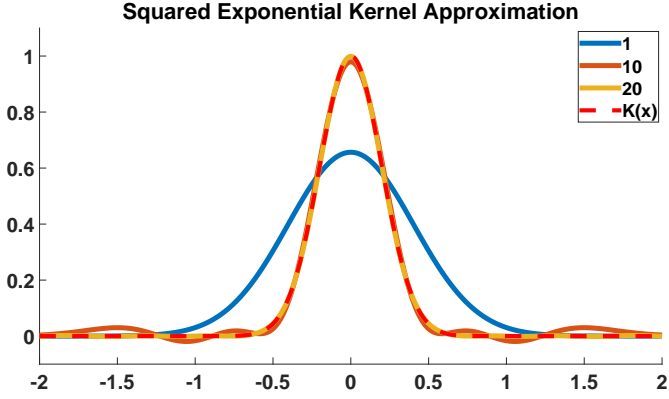}
    \caption{Approximation of the squared exponential kernel with $l_{se} = 0.2$ using 1, 10, and 20 eigen values. With just 20 eigenvalues the mean absolute difference between the approximation and the actual kernel values is \num{7.68e-4}.}
    \label{fig:exp_ker}
\end{figure}

The squared exponential covariance function 
\begin{align}
    \label{eq:se_ker}
    k_{se}(x,x') = e^\frac{-(x-x')^2}{2l_{se}^2}
\end{align}
is the most commonly used kernel in GP regression. It has a single hyperparameter $l_{se}$ which controls the kernel width and its Mercer expansion is given by \cite{fasshauer2012stable}:
\begin{align}
    \lambda_{se \ i} &= \sqrt{\frac{\alpha_{se}^2}{\alpha_{se}^2+\delta_{se}^2+\eta_{se}^2}}
    \left(\frac{\eta_{se}^2}{\alpha_{se}^2+\delta_{se}^2+\eta_{se}^2}\right)^i \\
    \Phi_{se \ i}(x) &= \sqrt{\frac{\beta_{se}}{i!}}e^{-\alpha_{se}^2x^2}H_i(\sqrt{2}\alpha_{se}\beta_{se}x) 
\end{align}
where $\eta_{se} = \frac{1}{\sqrt{2}l_{se}}$, $\beta_{se} = (1+(\frac{2\eta_{se}}{\alpha_{se}})^2)^{\frac{1}{4}}$, and $\delta_{se}^2 = \frac{\alpha_{se}^2}{2}(\beta_{se}^2-1)$. The parameter $\alpha_{se}$ is a tuning global scaling factor and can be utilized to avoid numerical issues with computing an inverse with extremely small eigenvalues. $H_i(\cdot)$ denotes the $i_{th}$ Hermite polynomial. The squared exponential kernel and its approximation using Mercer expansion are shown in figure \ref{fig:exp_ker}. 

One can interpret the expansion as a wavelet transform utilizing Hermitian wavelets. With this interpretation we see that the global scaling factor $\alpha_{se}$ in the eigen functions dilates or compresses the wavelet. Thus as the required range of $x$ increases one must decrease the scaling factor for the kernel approximation to maintain accuracy. Since $\alpha_{se}$ is also present in the eigenvalue equations, this in turn causes a slower eigenvalue drop off. Considering the scaling factor together with the kernel width parameter $l_{se}$ the implication is that one has to increase the number of eigenvalues for narrow kernels or when increasing the range of $x$. Finally, note that $l_{se}$ is present in both the eigen values and functions and thus for the squared exponential kernel the hyperparameter optimization requires re-evaluating $\mathbf{\Phi_X}$ at every iteration. 

\subsubsection{Squared Exponential Derivatives} \label{sec:exp_ker_der}
The $k_{th}$ derivative of $\Phi_{se \ i}(x)$ is calculated efficiently by noticing that $\frac{\partial^k e^{-\alpha_{se}^2x^2}}{\partial x^k}$ can be evaluated recursively
\begin{align}
    \nonumber
    \frac{\partial^k e^{-\alpha_{se}^2x^2}}{\partial x^k} &= P_k e^{-\alpha_{se}^2x^2} \\
    \nonumber
    P_0 &= 1, \ P_1 = -2\alpha_{se}^2x \\
    P_{k+1} &= -2\alpha_{se}(xP_{k} + (k-1)P_{k-1})
\end{align}
and $H_i(\sqrt{2}\alpha_{se}\beta_{se}x)$ represents an Appell sequence, thus
\begin{align}
    \frac{\partial^k H_i(\sqrt{2}\alpha_{se}\beta_{se}x)}{\partial x^k} = \frac{k!(\sqrt{2}\alpha_{se}\beta_{se})^k}{(i-k)!}H_{i-k}(\sqrt{2}\alpha_{se}\beta_{se}x) 
\end{align}
Finally applying Leibniz rule we obtain the $k_{th}$ derivative
\begin{align}
     \frac{\partial^k \Phi_{se \ i}(x)}{\partial x^k} = \sqrt{\frac{\beta_{se}}{i!}} \sum_{j=0}^k 
     \frac{\partial^{k-j} e^{-\alpha_{se}^2x^2}}{\partial x^{k-j}} \frac{\partial^j H_i(\sqrt{2}\alpha_{se}\beta_{se}x)}{\partial x^j}
\end{align} 

\subsubsection{Squared Exponential Hyperparameters}
Kernel length  $l_{se}$ is the only hyperparameter for this kernel, the gradient of $\lambda_{se \ i}$ with respect to the kernel length is a straightforward application of the chain rule. 
\begin{align}
    \nonumber
    \frac{\partial \lambda_{se \ i}}{\partial l_{se}} = &\biggr[ 2i\frac{\partial \eta_{se}}{\partial l_{se}}  + 
    \frac{\eta_{se}(-i-\frac{1}{2}) (\frac{\partial \delta^2_{se}}{\partial l_{se}} + 2\frac{\partial \eta_{se}}{\partial l_{se}}\eta_{se}) }{\alpha_{se}^2+\delta_{se}^2+\eta_{se}^2}\biggr] \\
    &\biggr[\alpha_{se} \eta_{se}^{2i-1}(\alpha_{se}^2+\delta_{se}^2+\eta_{se}^2)^{-i-\frac{1}{2}}  \biggr]
\end{align}
where
\begin{align}
    \frac{\partial \eta_{se}}{\partial l_{se}} &= -\frac{1}{\sqrt{2} l_{se}^2} \\
    \frac{\partial \beta_{se}}{\partial l_{se}} &= \frac{2}{\alpha_{se}^2}\frac{\partial \eta_{se}}{\partial l_{se}}\eta_{se}(1+(\frac{2\eta_{se}}{\alpha_{se}})^2)^{-\frac{3}{4}} \\
    \frac{\partial \delta^2_{se}}{\partial l_{se}} &= \alpha_{se}^2\frac{\partial \beta_{se}}{\partial l_{se}}\beta_{se}
\end{align}
Using chain rule and relying on the Appell sequence properties of $H_i(\cdot)$, the gradient of $\Phi_{se \ i}$ with respect to the kernel length $l_{se}$ can be evaluated efficiently as 
\begin{align}
    \nonumber
    \frac{\partial \Phi_{se \ i}(x)}{\partial l_{se}} &= \biggr( \frac{1}{2 \beta_{se}} \frac{\partial \beta_{se}}{\partial l_{se}}  -\frac{\partial \delta^2_{se}}{\partial l_{se}}x^2 \biggr) \Phi_{se \ i}(x) \\
    &+\sqrt{2i} \alpha_{se} \frac{\partial \beta_{se}}{\partial l_{se}} x \Phi_{se \ i-1}(x)
\end{align}

\subsection{Periodic Kernel}
\begin{figure}
    \centering
    \includegraphics[trim=5cm 1cm 5cm 0,clip,width=0.5\textwidth]{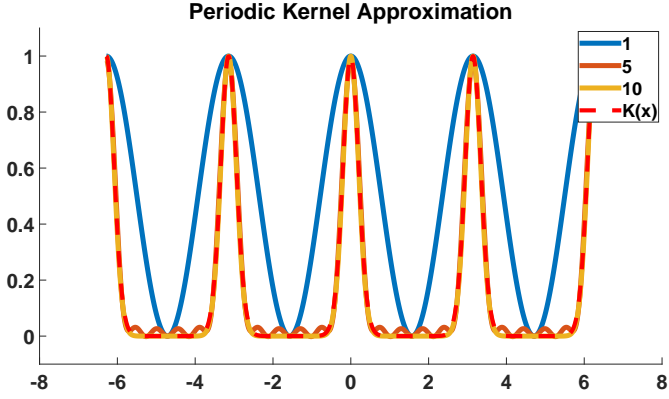}
    \caption{Approximation of the periodic kernel with $w_{pr} = 0.4$ and $f_{pr} = 2$ using 1, 5, and 10 eigen values. With only 10 eigenvalues the mean absolute difference between the approximation and the actual kernel values is \num{3.6e-3}.}
    \label{fig:per_ker}
\end{figure}

The periodic kernel covariance function 
\begin{align}
    k_{pr}(x,x') = e^{-\frac{2sin(f_{pr}\frac{(x-x')}{2})^2}{w_{pr}^2}}
\end{align}
allows to create Gaussian processes that are periodic. The frequency parameter $f_{pr}$ determines the distance between the repetitions and the width $w_{pr}$ controls the kernel width. The normalized Mercer expansion of the periodic kernel is derived in \cite{smola1998connection} and presents as a harmonic Fourier series. 
\begin{align}
    \nonumber
    \lambda_{pr \ 0} &= \frac{\gamma_{pr}}{\zeta_{pr}}, \ \ \Phi_{pr \ 0}(x) = 1 \\
     \lambda_{pr \ i} &= \begin{cases}
                \frac{e^{-\frac{j^2w^2}{2}}}{\zeta_{pr}} \ j=2i-1\\
               \frac{e^{-\frac{j^2w^2}{2}}}{\zeta_{pr}} \ j=2i
    \end{cases} \\
    \Phi_{pr \ i}(x) &= 
    \begin{cases}  
        cos(jf_{pr}x) \ \ \ j=2i-1\\
        sin(jf_{pr}x) \ \ \ j=2i
    \end{cases}
\end{align}
where $\gamma_{pr}$ and $\zeta_{pr}$ are the offset and scaling factor respectively to ensure the kernel has a range of $[0,1]$.
\begin{align}
    \gamma_{pr} &= \sum_{i=1}^n (-1)^{i-1}e^{-\frac{i^2w_{pr}^2}{2}}, \
    \zeta_{pr} = \sum_{i=1}^n 2e^{-\frac{(2i-1)^2w_{pr}^2}{2}}
\end{align}
The periodic kernel and its approximation are shown in figure \ref{fig:per_ker}.
The kernel width parameter $w_{pr}$ only appears in the eigen values, thus when learning a GP of a signal with a known period, $\mathbf{\Phi_X}$ does not need to be re-evaluated at every gradient descent iteration. One may also use this kernel for non-periodic signals by selecting $f_{pr}$ such that the kernel does not repeat in the range of $x$.

\subsubsection{Periodic Kernel Derivatives}
The sinusoidal structure of $\Phi_{pr \ i}(x)$ leads to easy evaluation of the $k_{th}$ derivative.
\begin{align}
    \frac{\partial^k \Phi_{pr \ i}(x)}{\partial x^k} = 
    \begin{cases}
        -(jf_{pr})^k sin(jf_{pr}x) \ \ \ j=2i-1\\
        (jf_{pr})^k cos(jf_{pr}x)  \ \  \ \ \ \ j=2i
    \end{cases}
\end{align}
Note that the above consists of scaled entries of $\Phi_{pr \ i}(x)$ and thus once $\mathbf{\Phi_X}$ is computed, $\frac{\partial^k \mathbf{\Phi_X}}{\partial \mathbf{x}^k}$ can be obtained directly.

\subsubsection{Periodic Kernel Hyperparameters}
The periodic kernel frequency $f_{pr}$ and width $w_{pr}$ parameters only appear in the eigen functions and values respectively. Utilizing the exponential and sinusoidal structures of the eigen values and functions the necessary derivatives for gradient descent parameter optimization are as follows:
\begin{align}
    \frac{\partial \Phi_{pr \ i}(x)}{\partial f_{pr}} &= 
    \begin{cases}
        -jx sin(jf_{pr}x) \ \ \ j=2i-1\\
        jx cos(jf_{pr}x)  \ \  \ \ \ \ j=2i
    \end{cases} \\
    \frac{\partial \lambda_{pr \ 0}}{\partial w_{pr}} &= \frac{1}{\zeta_{pr}}\frac{\partial \gamma_{pr}}{\partial w_{pr}}
    - \frac{\partial \zeta_{pr}}{\partial w_{pr}}\frac{\gamma_{pr}}{\zeta_{pr}^2} \\
    \frac{\partial \lambda_{pr \ i}}{\partial w_{pr}} &= -w_{pr}i^2\lambda_{pr \ i} - 
    \frac{\partial \zeta_{pr}}{\partial w_{pr}}\frac{\lambda_{pr \ i}}{\zeta_{pr}^2}
\end{align}
Where $\frac{\partial \gamma_{pr}}{\partial w_{pr}}$ and $\frac{\partial \zeta_{pr}}{\partial w_{pr}}$ are derivatives of the offset and scaling factors respectively. 
\begin{align}
\frac{\partial \gamma_{pr}}{\partial w_{pr}} &= \sum_{i=1}^n -(-1)^{i-1}w_{pr}i^2e^{-\frac{i^2w_{pr}^2}{2}}\\
\frac{\partial \zeta_{pr}}{\partial w_{pr}} &= \sum_{i=1}^n -2w_{pr}(2i-1)^2e^{-\frac{(2i-1)^2w_{pr}^2}{2} }
\end{align}

\subsection{Chebyshev Kernel}
\begin{figure}
    \centering
    \includegraphics[trim=5cm 1cm 5cm 0,clip,width=0.5\textwidth]{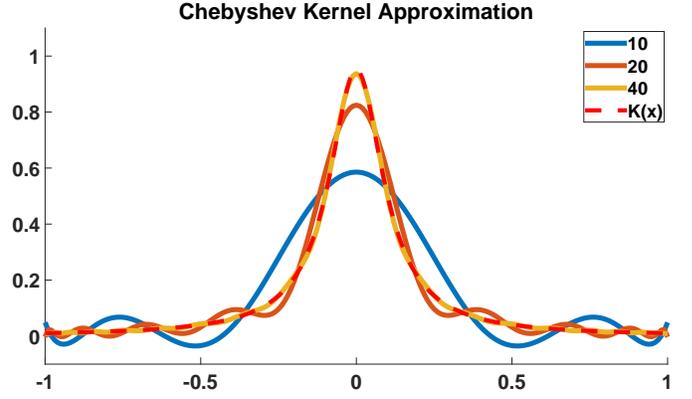}
    \caption{Approximation of the chebyshev kernel with $a = 0.9$ and $b = 0.9$ using 10, 20, and 40 eigen values.}
    \label{fig:che_ker}
\end{figure}

The final kernel function we include in this work is the Chebyshev kernel \cite{fasshauer2015kernel}.
\begin{align} 
    \label{eq:che_ker}
    k_{ch}(x,x') = 1-a\scriptstyle+\frac{2a(1-b)(b(1-b^2)-2b(x^2+x'^2)+(1+3b^2)xx')}
    {(1-b^2)^2 + 4b(b(x^2 + x'^2)-(1+b^2)xx')}
\end{align}
It has two hyperparameters $a \in (0, \ 1]$ and $b \in (0, \ 1)$ and a valid Mercer expansion in the range of $x \in [-1, \ 1]$.
\begin{align}
    \lambda_{ch \ 0} &= 1-a, \ \lambda_{ch \ i} = \frac{a(1-b)b^i}{b} \\
    \Phi_{ch \ 0}(x) &= 1, \ \Phi_{ch \ i}(x) = \sqrt{2}T_i(x)
\end{align}
where $T_i(\cdot)$ is the $i_{th}$ Chebyshev polynomial. The kernel and its approximation are illustrated in figure \ref{fig:che_ker}. Just like for the squared exponential kernel the expansion can be thought of as a wavelet transform, in this case using Chebyshev type wavelets. In our work this kernel function is of particular interest since all of the hyperparameters appear only in the eigen values. 

\subsubsection{Chebyshev Kernel Derivatives}
Similar to the Hermite polynomial derivatives presented in section \ref{sec:exp_ker_der}, the $k_{th}$ derivative of Chebyshev polynomial can be represented through Chebyshev polynomials of lower degrees \cite{prodinger2017representing}. 
\begin{align}
    \nonumber
    \frac{\partial^k T_i(x)}{\partial x^k} &= 2^k\sum_{j=0}^{(i-k)/2} i(i-1-j)^{\underline{k-1}} {k+j-1\choose k-1}T_{i-k-2j}(x) \\
    &-even(i-k)2^{k-1}n(\frac{i+k}{2}-1)^{\underline{k-1}}{\frac{i+k}{2}-1\choose k-1}
\end{align}
Where underlined superscript indicates falling factorials $x^{\underline{n}} = x(x-1)\dots(x-n+1)$ and the $even(\cdot)$ function outputs 1 for even arguments and 0 otherwise. Since $\Phi_{ch \ i}(x)$ is obtained by scaling $T_i(x)$, $\frac{\partial^k \mathbf{\Phi_X}}{\partial \mathbf{x}^k}$ can be computed efficiently from $\mathbf{\Phi_X}$ when using the Chebyshev kernel.

\subsubsection{Chebyshev Kernel Hyperparameters}
For this kernel the hyper parameters appear only in the eigen values allowing for extremely fast gradient descent based parameter optimization. 
\begin{align}
    \frac{\partial \lambda_{ch \ 0}}{\partial a} &= -1, \ \ \frac{\partial \lambda_{ch \ i}}{\partial a} = \frac{\lambda_{ch \ i}}{a} \\
    \frac{\partial \lambda_{ch \ 0}}{\partial b} &= 0, \ \ \frac{\partial \lambda_{ch \ i}}{\partial b} = -a(i(b-1)+1)b^{i-2}
\end{align}

It is important to note that for all of the presented expansions, as the width of the kernel decreases the number of eigenvalues necessary for an accurate approximation increases. Thus our method is particularly well suited when the number of data points is significantly larger than the number of eigen functions needed to accurately approximate the kernel. Using this approach with an inadequate number of eigen functions will lead to convergence to a wider kernel than optimal.

\section{Experiments} \label{sec:exp}
In this section we evaluate the computational complexity and accuracy of the proposed method. First we show that the training time of the proposed approach scales linearly when $\mathbf{\Phi_x}$ has to be re-evaluated every training iteration and is independent from the number of training points when the hyperparameters only appear in the eigen values. Next we evaluate the accuracy of the fast approximate multi-output GP considering both the numbers of training samples and eigenvalues. Finally, we show that the scaling matrix $K_f$ can correctly identify the correlation between outputs. 

\subsection{Computational Complexity}
\begin{figure}
    \centering
    \includegraphics[trim=4cm 0cm 4cm 0,clip,width=0.5\textwidth]{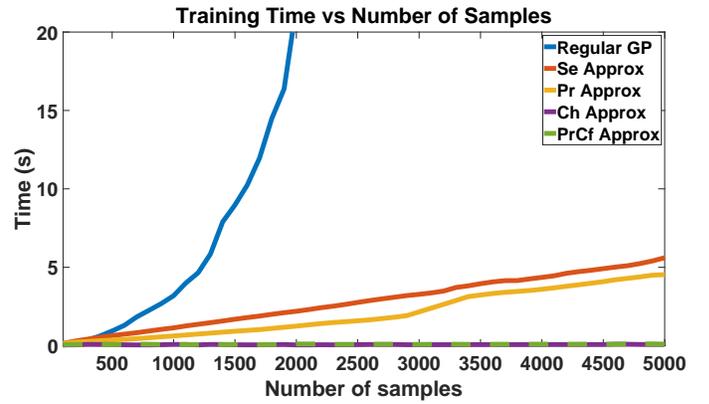}
    \caption{Required time to complete 100 iterations of gradient descent during hyperparameter optimization. Regular GP uses MATLAB's $fitgp$ function and we can observe the training time grow exponentially with the number of samples. When the proposed approach utilizes the Squared exponential (Se Approx) or Periodic (Pr Approx) kernel approximation it requires re-evaluating $\mathbf{\Phi_x}$ at every iteration and thus the training time is directly proportional to the number of samples. Employing the Chebyshev kernel (Ch Approx) or Periodic kernel with constant frequency (PrCf Approx) approximations requires only a single evaluation of $\mathbf{\Phi_x}$, during training the approach only updates the eigen values. For this demonstration the input $x$ is evenly spaced samples from (-1, 1) and the output is $sinc(x)$, 20 eigen values were used for all kernels. }
    \label{fig:itr_time}
\end{figure}

As discussed in section \ref{sec:famgp}, the proposed method requires only an inverse of $nM \times nM$ matrix instead of $NM \times NM$, where n, M, and N are the number of eigen values, outputs, and training samples respectively. During hyperparameter optimization the proposed approach further splits into two categories: (1) when the parameters are present in both eigen values and functions or (2) only in the eigen values. In the first case, $\mathbf{\Phi_x}$ needs to be re-evaluated after every training iteration, while in the second it is treated as constant and only the eigenvalues are updated. Figure \ref{fig:itr_time} shows the time it takes to complete 100 iterations of hyperparameter optimization using gradient descent for regular GP  and the two cases of the proposed approach. As expected, regular GP  quickly becomes intractable as the number of samples grows. In the proposed method, when $\mathbf{\Phi_x}$ needs to be re-evaluated at every iteration, the computational complexity grows linearly with the number of samples in the training set. When parameters are only present in the eigenvalues, the hypoerparameter learning time is independent from the number of samples in the training dataset.

\subsection{Accuracy}
\begin{figure}
    \centering
    \includegraphics[trim=3cm 0cm 5cm 0,clip,width=0.5\textwidth]{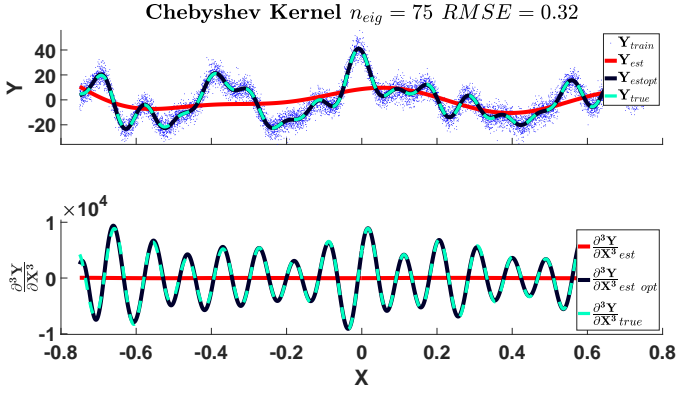}
    \caption{Regression of finite Fourier series with ZMG noise using FAMGP with the Chebyshev kernel approximation. In the top plot, blue dots and teal dashed line show the noisy training samples $Y_{train}$ and the noise free signal $Y_{true}$, red solid line $Y_{est}$ is the initial regression result before hyper parameter optimization $(a=0.5, \ b=0.5)$, black line $Y_{est \ opt}$ shows the regression after optimizing the hyper parameters $(a=0.998, \ b=0.954)$ using 5000 iterations of gradient descent which took 5.3 seconds to complete. The bottom plot shows the ability of the proposed approach to estimate the derivatives of the output, here we show estimated and actual jerk of the signal $(k=3)$.}
    \label{fig:che_ker_reg}
\end{figure}

To validate the regression accuracy we generate training data from an arbitrary generating function, using a sum of sinusoids of random frequencies, amplitudes, and phase shifts. This allows us to obtain the true $k_{th}$ derivative of the signal and verify that the proposed approach can correctly estimate high order derivatives. Zero mean Gaussian (ZMG) noise is added to the training data to simulate sensor noise.  The training data is generated from:

\begin{align}
\label{eq:sim_data}
    \mathbf{Y}_{true} = \sum_1^{10} c_i sin(f_i x + \varphi_i) + \epsilon_s
\end{align}
Where the amplitude coefficients $c_i$, frequencies $f_i$, and phase shifts $\varphi_i$ are drawn from a uniform distribution $U(1, \ 10)$ and $\epsilon_s \sim \mathcal{N}(0,5)$. The input variable $x$ consists of 10000 samples evenly spaced on the interval $[-5, \ 5]$. Figure \ref{fig:che_ker_reg} shows the regression capabilities of the Chebyshev kernel.

\begin{figure}
    \centering
    \includegraphics[trim=3cm 0cm 5cm 0,clip,width=\columnwidth]{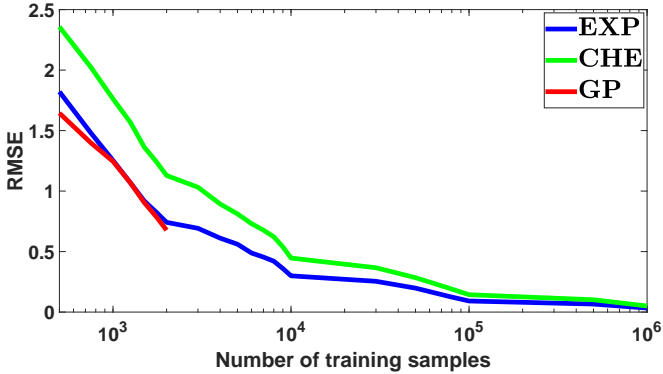}
    \caption{Regression accuracy improvement as the number of training samples increases. Regular GP training is not feasible for more than 2000 samples. FAMGP allows to optimize hyperparameters even with a million data points. The accuracy of both the squared exponential (EXP) and Chebyshev (CHE) kernel approximations converges as the number of samples increases. 75 eigenvalues were used for both kernels.}
    \label{fig:samples_rmse}
\end{figure}

As the number of training samples increases so should the regression accuracy. FAMGP allows us to utilize significantly larger training datasets. Figure \ref{fig:samples_rmse} shows the RMSE with respect to the number of training samples for regular GP, and FAMGP with squared exponential and Chebyshev kernels for the data presented in figure \ref{fig:che_ker_reg}. Due to computational complexity we are not able to utilize more than 2000 samples for the regular GP, FAMGP can easily be trained with a million, significantly improving the accuracy. The squared exponential kernel approximation provides lower RMSE compared to Chebyshev. However, Chebyshev kernel parameter optimization is significantly faster since $\mathbf{\Phi_x}$ is computed only once.

\iffalse
Table \ref{tab:reg_rmse} shows the regression accuracy if we increase the number of samples on the $[-5, \ 5]$ interval.
\begin{table}[]
\centering
\caption{Root mean squared error of the true signal $Y_{true}$ and proposed approach estimate $Y_{est \ opt}$ as the number of training samples on the $[-5, \ 5]$ interval increases. The Chebyshev kernel approximation is used with 75 eigen values. Without the proposed kernel approximation it would not be computationally feasible to optimize the hyper parameters while using one million training pairs.}
\begin{tabular}{|l|c|c|c|c|c|}
\hline
\textbf{Samples} & 100   & 1000  & \num{1e4} & \num{1e5} & \num{1e6} \\ \hline
\textbf{RMSE}             & 2.64 & 1.03 & 0.32 & 0.08  & 0.03   \\ \hline
\end{tabular}
\label{tab:reg_rmse}
\end{table}
\fi

\begin{figure}
    \centering
    \includegraphics[trim=3cm 0cm 5cm 0,clip,width=0.5\textwidth]{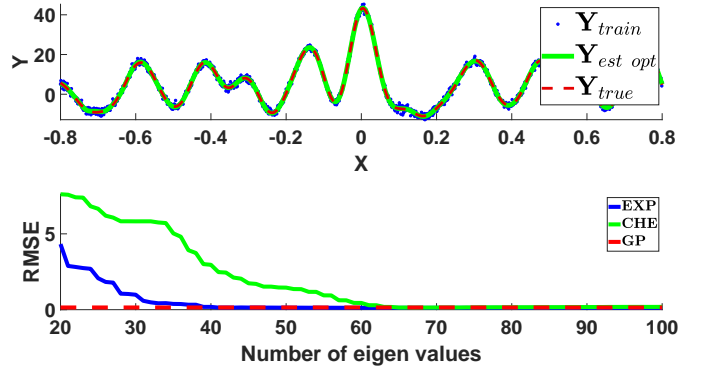}
    \caption{Regression RMSE as the number of eigenvalues increases. The top plot shows the noisy training data $\mathbf{Y}_{train}$, ground truth $\mathbf{Y}_{true}$, and FAMPG prediction $\mathbf{Y}_{est \ opt}$ signals when using the squared exponential kernel approximation with 100 eigenvalues. The bottom plot shows the regression RMSE as the number of eigenvalues increases from 20 to 100. The regression error of FAMGP converges to that of regular GP using both the squared exponential (EXP) and Chebyshev (CHE) kernel approximations. For the squared exponential kernel regular GP regression converges on width and scaling factors of $0.050$ and $225.55$ respectively, at 50 eigenvalues FAMGP optimization converged to very similar hyperparameter values $l_{se} = 0.048$ and $K_f = 215.22$, the eigenvalues sum to capture 97\% of data}
    \label{fig:rmse}
\end{figure}

Next we look at how the chosen number of eigenvalues effects the regression accuracy. We compare the performance of FAMGP with different number of eigenvalues to the standard GP formulation using the squared exponential kernel. Since the approximation can be interpreted as a wavelet transform, increasing the number of eigenvalues allows to accurately approximate a narrower kernel. Consider a sum of 10 sinusoids on the interval $x \in (-1, \ 1)$ with frequencies evenly distributed from $1$ to $10 rad/s$ and ZMG noise added of standard deviation of 0.1. Figure \ref{fig:rmse} shows the regression RMSE as we increase the number of eigenvalues. The accuracy and kernel parameters of the proposed approach converge to that of regular GP as the number of eigenvalues increases sufficiently to correctly approximate the narrow kernel. While the squared exponential kernel is the most commonly used covariance function when using GP regression, for FAMGP, the Chebyshev kernel is particularly attractive since $\mathbf{\Phi_x}$ does not need to be recalculated during hyperparameter optimization and allows for very fast training. The analysis shows that, while requiring more eigenvalues, the regression accuracy when using the Chebyshev kernel is comparable to that of squared exponential. 

\subsection{Correlation}
\begin{figure}
    \centering
    \includegraphics[trim=6cm 0cm 5cm 0,clip,width=\columnwidth]{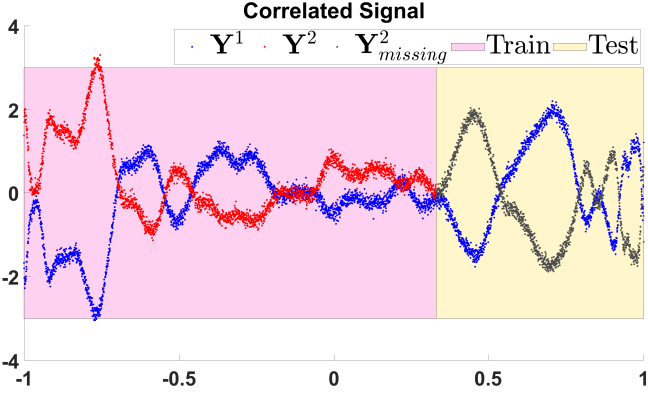}
    \caption{Two dimensional correlated training signal. The pink and yellow shaded regions are the training and test data sets respectively. Noisy output 1 ($Y^1$ \textcolor{blue}{blue}) is available both during training and testing. Output 2 is available for training ($Y^2$ \textcolor{red}{red}) but is missing from the test set ($Y^2_{missing}$ \textcolor{gray}{gray}).}
    \label{fig:corr_data}
\end{figure}

Finally we demonstrate that FAMGP  can correctly estimate the correlation between outputs and significantly improve regression when partial outputs are available. Furthermore, we compare the multioutput performance to that of regular GP \cite{bonilla2008multi} and show that both methods perform equally well and converge to almost identical correlation matrix and kernel parameters.  We sample 2000 training points of a highly correlated 2 dimensional signal from a zero mean normal distribution with a known covariance matrix generated utilizing equation \ref{eq:mv_cov}. The squared exponential covariance (eq. \ref{eq:se_ker}) with kernel parameters $l_{se}=0.1$ is used for $\mathbf{K_{XX}}$ and $x \in (-1, \ 1)$. High correlation between the outputs is achieved by setting $K_f$ as follows:
\begin{align}
    \nonumber 
    K_f = \begin{bmatrix}
    1.0 & -0.95\\
    -0.95 & 1.0
    \end{bmatrix}
\end{align}
Zero mean Gaussian noise is added to the output with $\Sigma_{Nm}=0.05 \mathbf{I}_{Nm}$. To test the ability of the proposed approach to utilize output correlation for regression we learn the kernel parameters and $K_f$ using the first 1333 data points. Next, $\mathbf{\alpha}'$ and $\mathbf{G}$ (table \ref{tab:prediction}) are computed utilizing all 2000 samples of output 1 and only the 1333 training samples of output 2. This simulates the situation where historical data of both correlated outputs is available for training. However, during regression, we have one output and would like to estimate the other. The data is visualized in figure \ref{fig:corr_data}.

\begin{figure}
    \centering
    \includegraphics[trim=6cm 0cm 5cm 0,clip,width=\columnwidth]{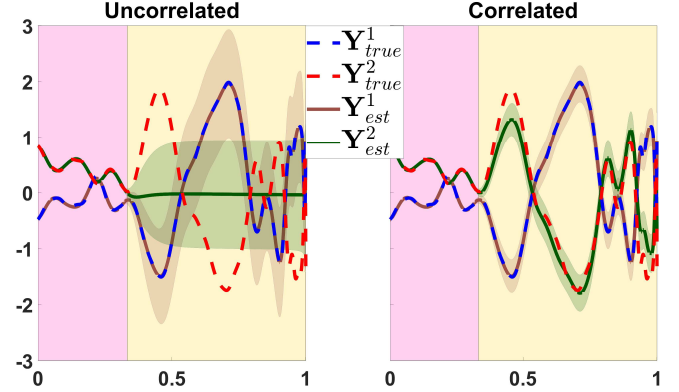}
    \caption{FAMGP regression over strongly correlated outputs. In the test region (yellow) noisy samples of $Y^1$ are available while $Y^2$ is entriely missing as explained in figure \ref{fig:corr_data}. Left: Uncorrelated output assumption, $K_f=\mathbf{I}_2$. When the outputs are assumed uncorrelated even though $Y^1$ is available for regression in the test region it is not utilized in estimation of $Y^2$ and the estimate drops to the zero mean assumption. Right: Using $K_f$ learned from the training region. Due to the correlation between outputs FAMGP can utilize the $Y^1$ samples in estimating $Y^2$ and maintain regression accuracy.}
    \label{fig:corr_uncorr}
\end{figure}

We train regular GP with the full squared exponential kernel and FAMGP with the kernel approximation utilizing 75 eigenvalues, initial kernel parameters of $l_{se}=0.5$ and initial correlation matrix set to identity, $K_f^{init} = \mathbf{I_2}$. Gradient descent converges on parameters shown in table \ref{tab:corr}. The method correctly estimates a strong negative correlation between the outputs even in the presence of significant noise. Figure \ref{fig:corr_uncorr} shows the FAMGP regression results over the test region when assuming independent outputs ($K_f=\mathbf{I}_M$) and using the learned correlation matrix, clearly demonstrating the benefits of the multivariate GP extension. Table \ref{tab:corr_rmse} compares the regression accuracy of FAMGP and regular GP for training and test data regions. Using 75 eigenvalues and functions to estimate a squared exponential kernel of length 0.1 is accurate to 99.99\% and thus the results between FAMGP and regular GP are almost identical. However, the training, regression, and storage requirements of FAMGP are magnitudes less than that of regular GP. For this example, at each training iteration FAMGP computes the $1333 \times 75 \ \mathbf{\Phi_X}$ matrix and evaluates a $150 \times 150$ inverse, regular GP calculates the full $1333 \times 1333$ kernel and the inverse of a $2666 \times 2666$ matrix. The proposed approach and regular GP took 45 and 441 seconds respectively to complete the required 926 gradient descent iterations for parameter convergence. After training, FAMGP needs to only save the $150$ element $\alpha'$ vector and $150 \times 150$ $\mathbf{G}$ matrix while GP needs the full $2666 \times 2666$ kernel inverse. Finally, for mean regression over the test set, FAMGP computes a $667 \times 75 \ \mathbf{\Phi_X}$ and multiplies it with the first 75 rows of $\alpha'$ to estimate $Y^1$ and last 75 rows for $Y^2$, GP requires $667 \times 1333$ kernel calculation and multiplication of the Kronecker product of the kernel and the correlation matrix with a $2666$ sized vector. The computational requirements grow linearly for FAMGP and exponentially for GP, thus while we can significantly increase the dataset size for the proposed approach, regular GP quickly becomes intractable.

\begin{table}[htbp]
  \centering
  \caption{Optimized $K_f$ matrix and kernel width for correlated outputs. Gradient descent converges to the true kernel width and accurately finds the negative correlation between outputs 1 and 2. The optimized parameters are very similar for both FAMGP and regular GP.}
    \hspace*{-6cm}\begin{tabular}{cll|l|@{\hspace{2em}}|l|l|l|}
    & & \multicolumn{2}{c}{FAMGP} & \multicolumn{2}{c}{GP} \\
    \multirow{3}[0]{*}{$K_f^{opt}=$}  &       & $Y^1$ & $Y^2$ & $Y^1$ & $Y^2$ \\ \cline{3-7}
          & $Y^1$ &  \cellcolor[rgb]{ .663,  .816,  .557}1.567 & \cellcolor[rgb]{ 1,  .486,  .502}-1.582 & \cellcolor[rgb]{ .663,  .816,  .557}1.559 & \cellcolor[rgb]{ 1,  .486,  .502}-1.554  \\ \cline{3-6}
          & $Y^2$  & \cellcolor[rgb]{ 1,  .486,  .502}-1.582  & \cellcolor[rgb]{ .663,  .816,  .557}1.706 & \cellcolor[rgb]{ 1,  .486,  .502}-1.554  & \cellcolor[rgb]{ .663,  .816,  .557}1.664 \\ \cline{3-6}
          $l_{se}^{opt}=$ & & \multicolumn{2}{c}{0.108} & \multicolumn{2}{c}{0.109}
    \end{tabular}%
  \label{tab:corr}%
\end{table}%

% Table generated by Excel2LaTeX from sheet 'Sheet1'
\begin{table}[htbp]
  \centering
  \caption{Regression root mean squared error for the correlated data split into training, test, and entire dataset.}
    \begin{tabular}{l|r|r|r|}
          & Train & Test & All \\ \hline
    FAMGP & 1.28E-04 & 0.0421 & 0.0141 \\ \hline
    GP    & 1.54E-04 & 0.0418 & 0.0141 \\ \hline
    \end{tabular}%
  \label{tab:corr_rmse}%
\end{table}%

\section{Conclusion and Future Work} \label{sec:fut}
In this work we presented a novel fast approximate multivariate Gaussian process framework. The key idea of the method is to approximate the covariance kernel using a finite number of eigenvalues and eigenfunctions. For a single output model this allows to reduce the required computational complexity of a GP training iteration from $\mathcal{O}(N^3)$ to $\mathcal{O}(n^3)$ where $N$ and $n$ are the number of training samples and eigenvalues respectively. In the multioutput case complexity is reduced from $\mathcal{O}((MN)^3)$ to $\mathcal{O}((Mn)^3)$ where M is the number of outputs. The proposed approach not only allows for fast training and estimation but also provides any order analytic derivatives of the GP. We provide the eigenvalues and functions of three different kernels (squared exponential, periodic, and Chebyshev) and show that in special cases hyperparameter optimization can be completely independent from the number of training samples. The method is extensively validated in simulation showing that depending on the optimal kernel width the proposed method's accuracy converges to that of regular GP with only a few eigenvalues. Our FAMGP implementation will be made publicly available \footnote{MATLAB source code will be available with the final submission at https://github.com/LucosidE/FAMGP}.

Currently the number of eigenvalues can be considered a tuning parameter of the algorithm, future work will include automatically increasing or reducing the number of eigenvalues during hyperparameter optimization by considering the ratio between the largest and smallest. This will allow training to speed up for wider kernels and maintain accuracy for very narrow ones. We also want to explore the applicability of the kernel approximations to multiple inputs, combining multiple kernels, and exploring additional available Mercer expansions. This would allow for learning much more complex processes. Finally, it may be possible to further optimize GP training and regression by combining the proposed approach with existing inducing points methods \cite{titsias2009variational, shi2020sparse} leading to Gaussian processes capable of handling extremely large datasets.

\bibliographystyle{IEEEtran}
\bibliography{main}

\end{document}